\documentclass[sigconf]{acmart}
\usepackage[ruled,vlined]{algorithm2e}

\AtBeginDocument{%
  \providecommand\BibTeX{{%
    \normalfont B\kern-0.5em{\scshape i\kern-0.25em b}\kern-0.8em\TeX}}}

\copyrightyear{2023} 
\acmYear{2023} 
\setcopyright{acmlicensed}\acmConference[SSDBM 2023]{35th International Conference on Scientific and Statistical Database Management}{July 10--12, 2023}{Los Angeles, CA, USA}
\acmBooktitle{35th International Conference on Scientific and Statistical Database Management (SSDBM 2023), July 10--12, 2023, Los Angeles, CA, USA}
\acmPrice{15.00}
\acmDOI{10.1145/3603719.3603731}
\acmISBN{979-8-4007-0746-9/23/07}

\begin{document}


\title{LearnedSort as a learning-augmented SampleSort: Analysis and Parallelization}

\author{Ivan Carvalho}

\orcid{0000-0002-8257-2103}
\affiliation{%
  \institution{University of British Columbia}
  \city{Kelowna}
  \country{Canada}
  \postcode{V1V 1V7}
}
\email{ivancarv@student.ubc.ca}

\author{Ramon Lawrence}
\orcid{0000-0002-6779-4461}
\affiliation{%
  \institution{University of British Columbia}
 \city{Kelowna}
  \country{Canada}
  \postcode{V1V 1V7}
}
\email{ramon.lawrence@ubc.ca}

\renewcommand{\shortauthors}{Ivan Carvalho and Ramon Lawrence}

\begin{abstract}
This work analyzes and parallelizes LearnedSort, the novel algorithm that sorts using machine learning models based on the cumulative distribution function. LearnedSort is analyzed under the lens of algorithms with predictions, and it is argued that LearnedSort is a learning-augmented SampleSort. A parallel LearnedSort algorithm is developed combining LearnedSort with the state-of-the-art SampleSort implementation, IPS4o. Benchmarks on synthetic and real-world datasets demonstrate improved parallel performance for parallel LearnedSort compared to IPS4o and other sorting algorithms.
\end{abstract}



\maketitle

\section{Introduction}

Sorting is a fundamental operation for databases and implementation details are relevant to sorting performance \cite{Graefe2006}. Quicksort is the asymptotically optimal comparison-based sort introduced by Hoare in the 1960s \cite{Hoare1962}. Its implementation variants have since become ubiquitous in database management systems such as PostgreSQL and standard libraries such as the GNU C++ library. Researchers keep pushing the boundary of sorting performance by proposing variants enhancing Quicksort \cite{Musser1997, Edelkamp2016, Peters2021, Axtmann2022, Wassenberg2022}.

LearnedSort is a novel sorting algorithm proposed by Kristo et al. \cite{Kristo2020} that introduces the idea of sorting with machine learning models. If there was a perfect model \(F(x)\) that predicted the position of \(x\) in the sorted array, we could sort the array \(A\) in a single pass with \(A[F(x)] = x\). Finding a perfect model is challenging, hence the LearnedSort authors opt to train cumulative distribution function (CDF) models to approximate \(F(x)\) and combine them with bucket partitioning. This approach by Kristo et al. was successful and led to excellent performance in sorting benchmarks. At first glance, LearnedSort looks radically different from Quicksort due to the CDF models. However, the analysis in this work demonstrates many similarities.

This work analyzes LearnedSort under the lens of algorithms augmented with predictions \cite{Mitzenmacher2022}. We highlight a close link between LearnedSort and SampleSort (a generalized version of Quicksort). This connection between the two algorithms has practical consequences. The first consequence is an explanation on why LearnedSort is so effective. The second consequence is the ability to leverage existing literature on SampleSort to address open issues of LearnedSort, such as sorting in parallel and handling inputs with many duplicates.

This paper makes the following contributions:

\begin{itemize}
\item
  Introduction of Learned Quicksort, a simpler version to analyze of LearnedSort
  that has average complexity of \(\mathcal{O}(N \log N)\)
\item
  Analysis showing that LearnedSort is analogous to a SampleSort with pivots selected by a CDF model
\item
  A new sorting algorithm combining the works of Kristo et al. \cite{Kristo2020} and Axtmann et al. \cite{Axtmann2022} to implement a parallel LearnedSort
\item
  Experimental benchmarking of the sorting implementations using the datasets from the LearnedSort paper \cite{Kristo2020}
\end{itemize}

\section{Background}

\subsection{Learned Indexes}

Learned Indexes are an emerging type of data structure that leverage machine learning models to create highly efficient indexes that outperform their traditional counterparts such as B-Trees \cite{Kraska2018}. Among such indexes, we highlight the Recursive Model Index (RMI) \cite{Kraska2018}, RadixSpline \cite{Kipf2020}, the Piecewise Geometric Model index  (PGM) \cite{Ferragina2020}, and the Practical Learned Index (PLEX) \cite{Stoian2021}.
Examples of scientific applications of learned indexes include geospatial indexing \cite{Pandey2020}, sensor time series indexing \cite{Ding2023}, and DNA sequencing \cite{Ho2019}.

Early work on learned indexes focused on in-memory, static key lookups. These indexes relied on approximating the CDF in order to find the key in a sorted array with few data accesses. The core idea is that if \(P(A \leq x)\) is the proportion of elements smaller than a key \(x\), then the position of the key \(x\) in a sorted array with \(N\) elements is \(pos = \lfloor N \times P(A \leq x) \rfloor\). 

Marcus et al. demonstrated that learned indexes were effective on read-only workloads \cite{Marcus2020_Benchmark}. Their experimental results show that the RMI, the PGM, and the RadixSpline always outperform state-of-the-art traditional indexes on look-up time and size, losing just on build time.

The two-layer RMI is used by LearnedSort. Mathematically, it is described by:
\[
F(x) = f_{2}^{\lfloor B \times f_{1}(x) \rfloor}(x)
\]

The RMI consists of the root model \(f_{1}\) and of \(B\) second-level models \(f_{2}^{(i)}\) for \(0 \leq i < B\). The root model can be interpreted as an initial approximation of the CDF function that selects one of the \(B\) models in the next level. The second-level models by consequence can be seen as models specializing on a specific region of the CDF.

The RMI architecture is extremely flexible. \(f_{1}\) and \(f_{2}^{(i)}\) can have arbitrary model types such as linear, cubic or radix models. The number of second-level models \(B\) can also be configured. LearnedSort uses a RMI with linear models and \(B = 1000\).

\subsection{Sorting with Machine Learning Models}

Sorting with machine learning models goes beyond applying a single pass of \(A[F(x)] = x\) for all elements. To engineer a practical implementation, many details need to be resolved.

The first detail is that \(A[F(x)] = x\) has a hostile memory-access pattern to modern CPUs, as it performs mostly random accesses to the memory. Kristo et al. reported that even with a perfect model, applying the model directly was slower than optimized versions of RadixSort. This prompted the authors to try other approaches such as using buckets.

Another key detail is that the model is imperfect and inversions can happen i.e. there are \(x, y\) such that \(x < y\) but \(F(x) > F(y)\). Although uncommon for good models, the implementation needs to handle those cases to guarantee that the output is sorted.

Moreover, collisions between elements can happen i.e. there are \(x, y\) such that \(F(x) = F(y)\). Since it is possible to have only one element at position \(F(x)\), the implementation must handle collisions. Collisions are exacerbated by duplicates in the input, as all duplicate values $x$ will collide at \(F(x)\). Duplicates are very common when sorting.

Kristo et al. improved the algorithm handling of these challenges and produced LearnedSort 2.0 \cite{Kristo2021}. LearnedSort 2.0 consists of four routines: training the model, two rounds of partitioning, model-based Counting Sort, and a correction step with Insertion Sort. 

Training the model is the first routine of LearnedSort and requires the most empirical data for good performance. It is necessary to select a model type and sample size to train the CDF model. Kristo et al. chose the two-layer RMI as the model. Since producing the optimal RMI is computationally more expensive than sorting an array with Quicksort \cite{Marcus2020_CDF}, the authors fixed the root and second-level model types to be linear models. They also picked a sample size of \(1 \%\) of \(N\) to train the RMI. 
These choices yield excellent results in practice. The model can be trained quickly and its predictions are accurate enough such that the sorting performance can outperform other state-of-the-art sorting algorithms.

The partitioning routine is in-place and uses the model to split the data into \(B = 1000\) buckets. For each element, LearnedSort calculates its corresponding bucket using \(b_{i} = \lfloor B  \times P(A \leq x) \rfloor\) and adds the element to the buffer associated with \(b_{i}\). When a buffer gets full, LearnedSort flushes the buffer. After processing all elements, the fragments of each bucket \(b_{i}\) are scattered across the input. To solve this, LearnedSort implements a defragmentation pass that makes the buckets contiguous. LearnedSort applies the partitioning routine twice, splitting the data into \(1000\) buckets and then splitting each of those buckets into \(1000\) sub-buckets.

To handle duplicates, Learned Sort 2.0 performs a homogeneity check after partitioning: if all elements within a bucket are equal, the bucket is left as is because it is already sorted. This condition handles the collision case that reduced the performance of the original LearnedSort.

The base case for LearnedSort is a Model-Based Counting Sort that uses the CDF to predict the final position of the keys in the sub-buckets. Lastly, Insertion Sort is executed to correct the possible mistakes from the RMI and guarantee that the output is sorted. Since the sequence is almost sorted, Insertion Sort is cheap to execute in practice.

\subsection{Quicksort}

Although Quicksort is asymptotically optimal, engineering a good implementation of Quicksort can drastically improve its efficiency. This requires avoiding Quicksort's worst case with bad pivots and squeezing all the performance available by modern hardware.

IntroSort is a hybrid Quicksort algorithm \cite{Musser1997} that avoids the \(\Theta (N^2)\) worst-case by switching to HeapSort \cite{Williams1964} when the recursion depth exceeds \(\mathcal{O}(\log N)\). IntroSort has been chosen by some popular libraries, such as the GNU C++ library, to be their default sorting algorithm.

Pattern-defeating Quicksort (pdqsort) is an enhanced version of IntroSort \cite{Peters2021}. It incorporates many improvements on partitioning and sorts in \(\mathcal{O}(N \min(\log N, K) )\) where \(K\) is the number of distinct elements on the input. pqdsort also leverages the contributions of BlockQuicksort \cite{Edelkamp2016}, which processes the elements in blocks to avoid branch mispredictions. pqdsort is currently the algorithm implemented by the Rust Standard Library for unstable sorting.

Vectorized Quicksort is a new implementation of Quicksort that uses Single Instruction, Multiple Data (SIMD) to exploit the parallelism available in modern CPUs \cite{Wassenberg2022}. Wassenberg et al. managed to vectorize each individual step of Quicksort: pivot selection, partitioning, and the sorting networks for the base case. By building on top of a high-level SIMD library, the authors were also able to port their implementation to seven distinct instruction sets, which is uncommon as previous implementations were generally not portable.

A takeaway from advancements in Quicksort is that engineering is a core part of high-performance sorting and that implementation details matter. Implementation optimizations improved performance in Learned Sort 2.0, and such optimizations are important for high parallel performance.

\subsection{SampleSort}

SampleSort is a generalization of Quicksort to \(k\) pivots \cite{Frazer1970}. The increased number of pivots pushes the number of comparisons of the algorithm closer to the \( \log_{2}n! \) theoretical bound, giving it an edge over Quicksort. It also makes the algorithm suitable for parallel processing, as SampleSort creates \(k + 1\) perfectly parallel sub-problems.

Similar to Quicksort, engineering a good implementation of SampleSort can significantly boost performance. Sanders and Winkel introduced the Super Scalar SampleSort in \cite{Sanders2004}. Their implementation of SampleSort exploits instruction-level parallelism available in modern CPUs. Sanders and Winkel organize the pivots into a branchless decision-tree that is friendly to optimization techniques such as pipelining and loop unrolling. This made their implementation competitive on single-core sequential settings.

Axtmann et al. take a step further in \cite{Axtmann2022}, introducing the In-place Parallel Super Scalar SampleSort  (\(\text{IPS}^{4}\text{o}\)). \(\text{IPS}^{4}\text{o}\) is the state-of-the-art SampleSort implementation incorporating many improvements.

One key improvement of \(\text{IPS}^{4}\text{o}\) is the in-place partitioning. Previous SampleSort implementations allocated \(\mathcal{O}(N)\) memory to copy elements of the input. \(\text{IPS}^{4}\text{o}\) instead uses buffers of size \(b\) for each of the \(k\) buckets. It allocates \(\mathcal{O}(kb)\) total memory
and when a buffer is full it flushes the buffer and overwrites some of the data of the original input that has already been processed. This initial pass creates \(\mathcal{O}(N/b)\) blocks. Afterwards, \(\text{IPS}^{4}\text{o}\) permutes the blocks such that each bucket is contiguous in memory using a routine similar to defragmentation. Conceptually, the blocking strategy adopted by \(\text{IPS}^{4}\text{o}\) shares many ideas with those adopted by LearnedSort, BlockQuicksort, and pdqsort.

Other improvements of \(\text{IPS}^{4}\text{o}\) include the parallelization and the equality buckets. \(\text{IPS}^{4}\text{o}\) uses atomic fetch-and-add operations to parallelize the block partitioning and leverages a custom task scheduler to manage threads when the sub-problems become small. \(\text{IPS}^{4}\text{o}\) also gracefully handles inputs with many duplicates with equality buckets. It detects skewed inputs on sampling and creates a separate bucket for the duplicates when doing the partitioning. As a sequence where all elements are equal is already sorted, \(\text{IPS}^{4}\text{o}\) avoids having to process the duplicate elements in the equality buckets. 

It is also worth highlighting the ability to use \(\text{IPS}^{4}\text{o}\) as a framework for building other sorting algorithms. Axtmann et al. also introduced the In-place Parallel Super Scalar Radix Sort (\(\text{IPS}^{2}\text{Ra}\)) \cite{Axtmann2022}. \(\text{IPS}^{2}\text{Ra}\) combines the qualities of \(\text{IPS}^{4}\text{o}\) with the most-significant-digit radix sort strategy, resulting in another high-performance sorting algorithm. \(\text{IPS}^{4}\text{o}\) has also been used to parallelize Vectorized Quicksort \cite{Wassenberg2022} and to test the efficiency of sorting networks as base cases for sorting algorithms \cite{Bingmann2020}.

This work reuses the \(\text{IPS}^{4}\text{o}\) framework to parallelize LearnedSort. This allows the combination of the engineering efforts of \(\text{IPS}^{4}\text{o}\) with the best qualities of LearnedSort. 

\subsection{Algorithms with Predictions}

The area of algorithms with predictions \cite{Mitzenmacher2022} goes beyond worst-case analysis and considers algorithms augmented with machine learning models. For each algorithm, we can think of a prediction and a quality metric \(\eta\) for the prediction that depends on an error specified by the problem type. In case \(\eta\) is good, the algorithm proceeds to use the outputs from the model to solve the problem instance. Otherwise, it has a fallback mechanism that uses a traditional, prediction-less algorithm when the machine learning models fail. We expect that for real-world workflows, the outputs from the model will generally be used due to patterns found in the data.

A prominent example of the area is caching with predictions \cite{Lykouris2021}. Lykouris and Vassilvitskii solve the online caching problem with a machine learning model trained to predict the furthest time in the future an element will come back to the cache. Their model is inspired by the offline solution to the problem, the greedy Furthest-In-Future algorithm that out of all elements removes the one that appears the latest in the future. To prevent the worst-case that happens when the model is sub-optimal, they fall back to the classic Marker algorithm.

Algorithms with predictions share many similarities with LearnedSort. Both implement machine learning models and avoid the worst-case due to the quality of the predictions. Thus, it is natural to ask if LearnedSort is an algorithm with predictions. 
The next section discusses how LearnedSort is analogous to a SampleSort in which the pivots were learned.

\section{Analyzing LearnedSort}

To analyze LearnedSort under the lens of algorithms with predictions, it is important to determine what LearnedSort is trying to predict and what makes for a good prediction for a sorting algorithm.

From a high-level perspective, ignoring implementation details, what makes Quicksort an efficient algorithm is the quality of its pivots. The BFPRT algorithm, also known as median of medians, is a method to find an element that is guaranteed to be between the 30th and 70th percentile of the input \cite{Blum1973}. It is possible to combine Quicksort with the BFPRT to produce a deterministic Quicksort with worst-case complexity of \( \Theta(N \log N) \) \cite{Kurosawa2016}. Hence, the quality of the pivots can avoid the worst-case of randomized Quicksort. 

Inspired by the deterministic Quicksort, the analysis of LearnedSort in split into three parts. The first part introduces Quicksort with Learned Pivots, a variation of Quicksort where the CDF model selects the pivot. That section shows that training a CDF model is akin to other pivot selection techniques such as applying the BFPRT algorithm. The second part analyzes Learned Quicksort, a simplified LearnedSort with \(B = 2\) buckets. It turns out that Learned Quicksort is in fact analogous to a Quicksort with Learned Pivots but with implicit pivots. Lastly, the third section considers \(B > 2\) and the connections between LearnedSort and SampleSort.

\subsection{Quicksort with Learned Pivots}

The analysis starts with the pseudocode of our Quicksort variant shown in Algorithm \ref{alg:Quicksort}. For simplicity, assume that all elements on the input \(A\) are distinct. The algorithm is identical to many other Quicksort implementations with the exception of the partitioning call.

\begin{algorithm}
\caption{Quicksort(A, l, r)}\label{alg:Quicksort}

\If{distance(l, r) \(\leq\) BASECASE\_SIZE}{
   InsertionSort(A, l, r);
   
   \Return{};
}

q \(\gets \) PartitionWithLearnedPivot(A, l, r);

Quicksort(A, l, q-1);

Quicksort(A, q + 1, r);

 \Return{};
\end{algorithm}

Algorithm \ref{alg:PartitioningQuicksort} describes how to use the CDF models to select an optimal pivot. Essentially, our goal is to find the median of the input. To do so, we select the largest element \(A[t]\) such that the predicted CDF is smaller than or equal to the true CDF of the median.

\begin{algorithm}
\caption{PartitionWithLearnedPivot(A, l, r)}\label{alg:PartitioningQuicksort}

S \(\gets \) Sample(A, l, r);

HeapSort(S, 0, S.size() - 1);

F \(\gets \) TrainCDFModel(S, 0, S.size() - 1); \tcp{Function that calculates P(A <= x) in [0, 1]}

 \tcc{Select the largest element from A that has the predicted CDF less than the true median}

t \( \gets\) -1;

 \For{\(w \gets l\) \textbf{to} \(r\)}{  
  \If{F(A[w]) \(\leq\) 0.5 and (t < 0 or A[w] > A[t])  }{
   \(t \gets\) w;
   }
 }

swap(A[t], A[r]);

 \tcc{After selecting the pivot with the CDF model, we can use any classic partition scheme}

pivot \( \gets \) A[r];

i \( \gets\) l - 1;

 \For{\(j \gets l\) \textbf{to} \(r - 1\)}{  
  \If{A[j] \(\leq\) pivot }{
   
   i \( \gets\) i + 1;

   swap(A[i], A[j]);
   }
 }

 swap(A[i], A[r]);
 
 \Return{i + 1};
\end{algorithm}

The TrainCDFModel function is arbitrary such that any type of CDF model could work e.g. RMI, PLEX, RadixSpline. However, for the CDF model to be useful, some properties should hold. 

The first is monotonicity: \(x \leq y \implies F(x) \leq F(y)\). This property is necessary to ensure that the selected pivot is indeed closest to the median and that the model contains no incorrect inversions. 

The second is that the model needs to require a small number of samples. This follows from the fact that to train a CDF model you need a sorted input and sorting the samples with HeapSort takes \(\mathcal{O}(S \log S)\) (although any algorithm with the same complexity would work). 

The third is that computing the predictions of the model for a key should take \(\mathcal{O}(1)\) time. Since we need to make a prediction for each of the \(N\) keys, if the time to compute a prediction is not constant it would lead to an algorithm slower than the traditional Quicksort.

Given these properties, Algorithm \ref{alg:PartitioningQuicksort} takes \(\mathcal{O}(N)\) and its run time is dominated by the loop applying the model predictions and the Lomuto partitioning step.

The time complexity of Algorithm \ref{alg:Quicksort} depends on the quality of the learned pivot. In the best case, the complexity is modelled by \(T(N) = \mathcal{O}(N) + 2 T(N/2)\) which happens when the learned pivot is the median. Hence, the lower bound of Algorithm \ref{alg:Quicksort} is \(\Omega (N \log N)\).

The worst-case complexity is modelled by \(T(N) = \mathcal{O}(N) + T(N - 1)\) and happens when the learned pivot is the smallest element in the sequence. Thus, the worst-case of the algorithm is \(\Theta (N^2)\)  just like the original Quicksort. However, if the chosen model is a good model, reaching the worst-case is unlikely. The average-case analysis is much closer to the best case in practice.

Let \(\eta\) be the error from finding the perfect partitioning as:

\[\eta = \max( P(A \leq pivot), 1 - P(A \leq pivot) ) - 1/2  \]

where \(P(A \leq pivot)\) is the true CDF of the learned pivot. \(\eta = 0\) in case the CDF model always predicts the median. \(\eta = 1/2\) in case the CDF model always predicts the smallest element. The complexity is then modelled by: 

\[T(N) = \mathcal{O}(N) +  T( (\eta + 1/2) N) +  T( (-\eta + 1/2) N)\] 

The value \( \eta \) is not known ahead of time, as it depends on the sample size and CDF model. However, we may assume that the model has better predictions than a random pick \( \eta_{\text{learned}} \leq \eta_{\text{random}}\) (otherwise we would fallback to a random pick). This implies that the Quicksort with Learned Pivots runs as fast as the Randomized Quicksort. Thus \( T(N) \in \mathcal{O} (N \log N) \).

Quicksort with Learned Pivots is not efficient to outperform IntroSort or pdqsort. However, it is conceptually useful to show that training a CDF model is a step towards finding better pivots.

\subsection{Learned Quicksort}

Progressing towards analyzing LearnedSort, we introduce Learned Quicksort. Learned Quicksort, shown in Algorithm \ref{alg:LearnedQuicksort}, is a simpler version of LearnedSort that contains only \(B = 2\) buckets.

\begin{algorithm}
\caption{LearnedQuicksort(A, l, r)}\label{alg:LearnedQuicksort}

\If{distance(l, r) \(\leq\) BASECASE\_SIZE}{
   InsertionSort(A, l, r);
   
   \Return{};
}

S \(\gets \) Sample(A, l, r);

HeapSort(S, 0, S.size() - 1);

F \(\gets \) TrainCDFModel(S, 0, S.size() - 1);

\tcc{Using the predictions directly is equivalent to using the learned pivot}

i \(\gets \) l;

j \(\gets \) r;

\While{i < j}{
\eIf{F(A[i]) \(\leq\)  0.5}{

i \(\gets\) i + 1;

}{
swap(A[i], A[j]);

j \(\gets\) j - 1;
}
}

LearnedQuicksort(A, l, i);

LearnedQuicksort(A, i + 1, r);

 \Return{};
\end{algorithm}

Similar to LearnedSort, Learned Quicksort partitions the data using machine learning models. Since there are only two buckets, the partitioning can be done such that elements with \(F(A[i]) \leq 0.5\) are put in the initial section of the input starting from the first index and elements with \(F(A[i]) > 0.5\) are put at the end of the input starting from the last index.

The partitioning done by Quicksort with Learned Pivots and Learned Quicksort is almost identical. The only exception is for the learned pivot, which is in the last position of the first half in the former. Hence, the algorithms have the same time complexity which means that Learned Quicksort has the complexity of \( \mathcal{O} (N \log N) \).

The interesting fact about Learned Quicksort is that it does not compute the pivot explicitly. Instead, it relies solely on the results of the model \(F\). Computationally, this is advantageous as Learned Quicksort always performs less operations than Quicksort with Learned Pivots.

We may interpret Learned Quicksort as a Quicksort variant that circumvents the bounds on the theoretical number of comparisons by embracing the numerical properties of the CDF. This is a hint to why LearnedSort is so efficient.

\subsection{LearnedSort}

We now consider the general case of LearnedSort when \( B > 2 \). If Learned Quicksort is analogous to a Quicksort with Learned Pivots, LearnedSort is analogous to a SampleSort with \(B - 1\) learned pivots.

\begin{algorithm}
\caption{LearnedPivotsForSampleSort(A, l, r)}\label{alg:LearnedPivotsForSampleSort}

B \(\gets\) numberOfBuckets(A.size());

p \(\gets\) Array(B, -1); \tcp{indexes of the pivots for the i-th bucket} 

S \(\gets \) Sample(A, l, r);

HeapSort(S, 0, S.size() - 1);

F \(\gets \) TrainCDFModel(S, 0, S.size() - 1);

 \tcc{Select for each i-th percentile the largest element from A that has predicted CDF less than that percentile}

 \For{\(w \gets l\) \textbf{to} \(r\)}{
  g \( \gets \lfloor F(A[w]) \times B  \rfloor \) ;

  \If{p[g] < 0 or  A[w] > A[p[g]] }{
   p[g] \( \gets\) w;
   }
 }

pivots \(\gets\) p.filter(v \( \geq 0\)).map(v \(\rightarrow\) A[v]);
 
 \Return{pivots};
\end{algorithm}

Algorithm \ref{alg:LearnedPivotsForSampleSort} details the process to compute the learned pivots for SampleSort. If we used those pivots in SampleSort, the partitioning would be identical to the partitioning done by LearnedSort. Obviously, as shown in the previous section, using the model directly is more advantageous as it skips the comparisons as a whole.

This explains why LearnedSort is effective. LearnedSort is an enhanced version of SampleSort, which is already a competitive sorting algorithm. If the learned pivots of LearnedSort are better than the randomly selected pivots of SampleSort, we expect LearnedSort to beat SampleSort. Moreover, LearnedSort skips the comparisons done by SampleSort and relies on the \( \mathcal{O}(1) \) predictions of the model, which gives LearnedSort an additional performance boost.

There are some differences between an augmented SampleSort and the implementation of LearnedSort 2.0. These minor details come from the authors iterating to improve LearnedSort empirically.

The major discrepancy is that SampleSort does \( \mathcal{O}( \log_{B} N ) \) partitioning rounds while LearnedSort does only two. We interpret this as Kristo et al. implementing a very shallow recursion tree with a large base case size. SampleSort implementations generally use \(B = 128\) or \(B = 256\) buckets and use Insertion Sort when there are 16 elements or less. LearnedSort uses \(B = 1000\) buckets, hence assuming two rounds of partitioning with around \(N = 10^9\) elements of input that leads to around 1000 elements on average to be handled by LearnedSort's base case. We hypothesize that if there were \(N = 10^{12}\) or \(N = 10^{13}\) elements, LearnedSort's performance would be hurt and a third partitioning round would be helpful. However, that input size requires over a terabyte of RAM, which stops being an in-memory sort problem and starts being an external sort instance. Thus, the implementation by Kristo et al. works well in practice.

Another discrepancy is that SampleSort samples data on every sub-problem while LearnedSort samples data only once. This may be an optimization that comes from practical experience. Instead of sampling a few data points, creating 1000 sub-problems and sampling for each sub-problem again, LearnedSort opts to sample a lot of data in bulk. This works because the recursion tree of LearnedSort is very shallow and because the RMI architecture supports this strategy as the second-level models specialize in parts of the CDF.

Lastly, the RMI used by LearnedSort violates one assumption from our analysis. It does not guarantee that \(x \leq y \implies F(x) \leq F(y)\). In practice, inversions do occur but they are relatively rare. This leads to an almost-sorted sequence, which can be quickly fixed by Insertion Sort.

\subsection{Quality of the Pivots}

This section analyzes the quality of the learned pivots implicitly used by LearnedSort. For two datasets, the uniform distribution and the Wiki/Edit data, the RMI created by LearnedSort was used with Algorithm \ref{alg:LearnedPivotsForSampleSort} to calculate the pivots in the first partitioning step. The RMI pivots were compared with the random pivots used by  \(\text{IPS}^{4}\text{o}\).

\begin{table}[H]
\caption{Comparison of Pivot Quality}
\begin{tabular}{|l|l|l|}
\hline
          & Random (255 pivots) & RMI (255 pivots) \\ \hline
Uniform   & 1.1016              & 0.4388                   \\ \hline
Wiki/Edit & 0.9991              & 0.5157                   \\ \hline
\end{tabular}
\caption{Quality of the pivots for $\text{IPS}^{4}\text{o}$ (Random) and LearnedSort (RMI)} \label{tab:Quality} 
\end{table}

The sorted data was used to calculate the true CDF, \( P(A \leq p_{i}) \), for each pivot \(p_{i}\). The metric used for the quality was the distance between the CDF of the pivots and the CDF of the perfect splitters  \( \sum_{i=0}^{B - 2} |P(A \leq p_{i}) - (i+1)/B| \). For simplicity, we matched the number of pivots used by \(\text{IPS}^{4}\text{o}\) with the number of pivots computed by the RMI, although LearnedSort uses more pivots in practice.

The results in Table \ref{tab:Quality} display that the learned pivots are indeed better than the random pivots.

\section{Parallelization of LearnedSort}

One direct consequence from the previous analysis is that the progress in engineering a fast SampleSort transfers to LearnedSort. A relevant limitation of LearnedSort 2.0 is that there is only a sequential version available that cannot use all the cores present in modern CPUs to sort data in parallel. This limits applying LearnedSort to real-world workflows.

To address this limitation, we introduce the Augmented In-place Parallel SampleSort (\(\text{AIPS}^{2}\text{o}\)). \(\text{AIPS}^{2}\text{o}\) is a hybrid of \(\text{IPS}^{4}\text{o}\) with LearnedSort. It is built upon the codebase available from \(\text{IPS}^{4}\text{o}\) and augments it with the RMI implementation used in LearnedSort.

\begin{algorithm}
\caption{BuildPartitionModel(A, l, r)}\label{alg:BuildPartitionModel}

S \(\gets \) Sample(A, l, r);

Sort(S, 0, S.size() - 1);

\eIf{InputSizeIsLarge(l, r) and not TooManyDuplicates(S) }{
    \tcp{we sample more data as the RMI benefits from larger samples} 
   
   R \(\gets \) LargerSample(A, l, r);

    Sort(R, 0, R.size() - 1);
   
   rmi \( \gets \) BuildRMI(R);

   \Return{rmi};
   } {
   tree \( \gets \) BuildBranchlessDecisionTree(S);

   \Return{tree};
   }

\end{algorithm}

How \(\text{AIPS}^{2}\text{o}\) selects its partitioning strategy is in Algorithm \ref{alg:BuildPartitionModel}. Essentially, if the input size is sufficiently large and there are not too many duplicates, the routine samples more data and returns a trained RMI. Otherwise, it builds and returns the decision tree used in \(\text{IPS}^{4}\text{o}\). For our implementation, we use \(B = 1024\) buckets for the RMI. We default to the decision tree with \(B = 256\) if the input size is smaller than \(N = 10^{5}\) or if there are more than 10\% of duplicates in the first sample.

Since \(\text{AIPS}^{2}\text{o}\) uses the framework from \(\text{IPS}^{4}\text{o}\), it profits from the parallelization of the latter. Another feature it inherits from \(\text{IPS}^{4}\text{o}\) is the handling of duplicates, which avoids the common adversarial case for LearnedSort by using the equality buckets from the decision tree.

There are additional modifications to make \(\text{AIPS}^{2}\text{o}\) work as well. The most critical modification is making the RMI monotonic such that \(x \leq y \implies F(x) \leq F(y) \) holds. This is necessary to avoid applying Insertion Sort to guarantee the correctness. To implement a monotonic RMI, we had to constraint the second-level linear models such that \( \max_{x \in R} f_{2}^{(i)}(x) \leq \min_{x \in R} f_{2}^{(i+1)}(x) \). This incurs two additional accesses to an array storing the minimums and maximums when processing an element.

The base case is also modified. Model-based counting sort is not used as the algorithm never forwards the RMI between recursive calls. Instead, SkaSort is used for the base case when there are less than 4096 elements \cite{skasort2016}. SkaSort is a fast radix sort that is the base case for \(\text{IPS}^{2}\text{Ra}\).

\section{Experimental Results}

\(\text{AIPS}^{2}\text{o}\) is compared against other sorting algorithms
on the benchmark presented in the Learned Sort 2.0 paper \cite{Kristo2021}. For reproducibility, benchmarks were executed on the \textbf{m5zn.metal} instance from AWS. The instance runs an Intel® Xeon® Platinum 8252C CPU
@ 3.80GHz with 48 cores, 768KB of L1 cache, 24MB of L2 cache, 99 MB of L3 cache, and 192 GB
of RAM.

The four competitors with \(\text{AIPS}^{2}\text{o}\) are the following. \(\text{IPS}^{4}\text{o}\), the state-of-the-art implementation of SampleSort. \(\text{IPS}^{2}\text{Ra}\), the radix sort implementation built on top of the framework for \(\text{IPS}^{4}\text{o}\). Learned Sort, one of the fastest sequential sorting algorithms as discussed earlier. \texttt{std::sort} from the C++ STL, as the baseline for the experiment. The implementations were written in C++ and compiled with GCC 11 using the -O3 and -march=native flags.

The benchmark includes sequential and parallel settings. We refer to the sequential versions of the algorithms  as \(\text{AI1S}^{2}\text{o}\), \(\text{I1S}^{4}\text{o}\), and \(\text{I1S}^{2}\text{Ra}\) for consistency as they are not parallel. We provide \texttt{std::execution::par\_unseq} as an argument to \texttt{std::sort} when executing in parallel. To sort floating point numbers with \(\text{IPS}^{2}\text{Ra}\), we use a key extractor that maps floats to integers. Learned Sort is not in the parallel benchmark because there is only a sequential implementation. The parallel benchmark uses all of the 48 cores available in the machine.

The datasets used in the benchmark consist of synthetic and real-world data. The synthetic portion contains 64-bit double floating-point elements from various probability distributions. The real-world portion contains 64-bit unsigned integer elements mostly from the work of Marcus et al. \cite{Marcus2020_Benchmark}. For {\bf N} = \(10^{8}\), data size is 800 MB.

\textbf{Synthetic Datasets}

\begin{itemize}
\item
  \textbf{Uniform (N = \(10^{8}\))}: Uniform distribution with \(a=0\) and
  \(b=N\)
\item
  \textbf{Normal (N = \(10^{8}\))}: Normal distribution with \(\mu=0\) and
  \(\sigma=1\)
\item
  \textbf{Log-Normal (N = \(10^{8}\))}: Log-normal distribution with \(\mu=0\) and
  \(\sigma=0.5\)
\item
  \textbf{Mix Gauss (N = \(10^{8}\))}: Random additive
  distribution of five Gaussian distributions
\item
  \textbf{Exponential (N = \(10^{8}\))}: Exponential Distribution with \(\lambda = 2\)
\item
  \textbf{Chi-Squared (N = \(10^{8}\))}: \(\chi ^ 2\) distribution with \(k = 4\)
\item
  \textbf{Root Dups (N = \(10^{8}\))}: Sequence of
  \(A[i] = i \mod \sqrt{N}\) as proposed in \cite{Edelkamp2016}
\item
  \textbf{Two Dups (N = \(10^{8}\))}: Sequence of
  \(A[i] = i^2 + N / 2 \mod N\) as proposed in \cite{Edelkamp2016}
\item
  \textbf{Zipf (N = \(10^{8}\))}: Zipfian distribution
  with \(s_{\text{zipf}} = 0.75\)
\end{itemize}

\textbf{Real-World Datasets}

\begin{itemize}
\item
  \textbf{OSM/Cell\_IDs (N = \(2 \cdot 10^{8}\))}: Uniformly sampled
  location IDs from OpenStreetMaps.
\item
  \textbf{Wiki/Edit (N = \(2 \cdot 10^{8}\))}: The edit timestamps from
  Wikipedia articles
\item
  \textbf{FB/IDs (N = \(2 \cdot 10^{8}\))}: The IDs from Facebook users
  sampled in a random walk of the network graph
\item
  \textbf{Books/Sales (N = \(2 \cdot 10^{8}\))}: Book popularity data
  from Amazon
\item
  \textbf{NYC/Pickup (N = \(10^{8}\))} : The yellow taxi trip pick-up
  time stamps
\end{itemize}

\subsection{Sequential Results}

\begin{figure}[!htbp]
    \includegraphics[width=\linewidth]{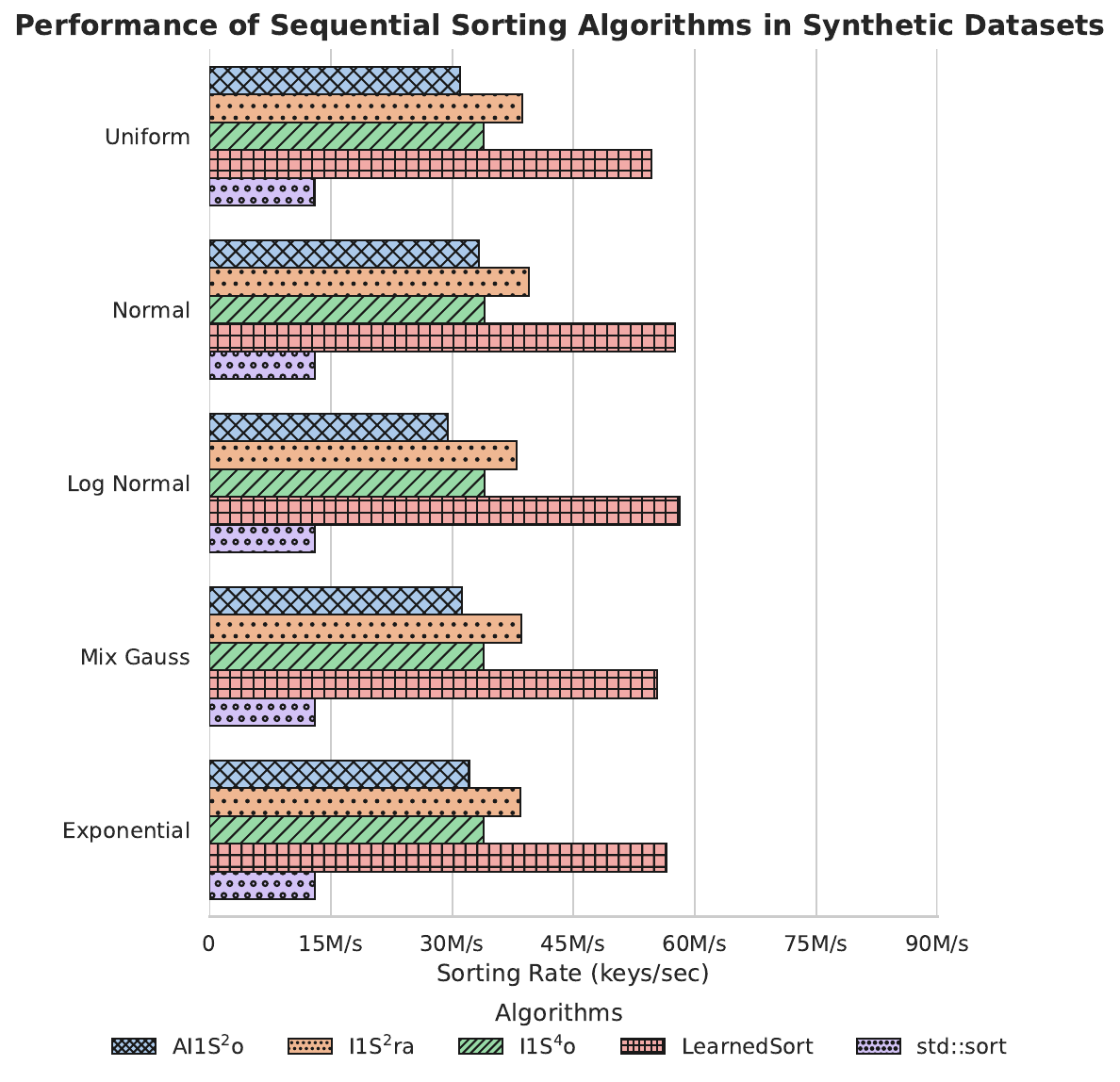}
    \caption{Sorting throughput of the sequential algorithms. Higher rates are better.}
    \label{fig:seq_synth_a}
\end{figure}

\begin{figure}[!htbp]
    \includegraphics[width=\linewidth]{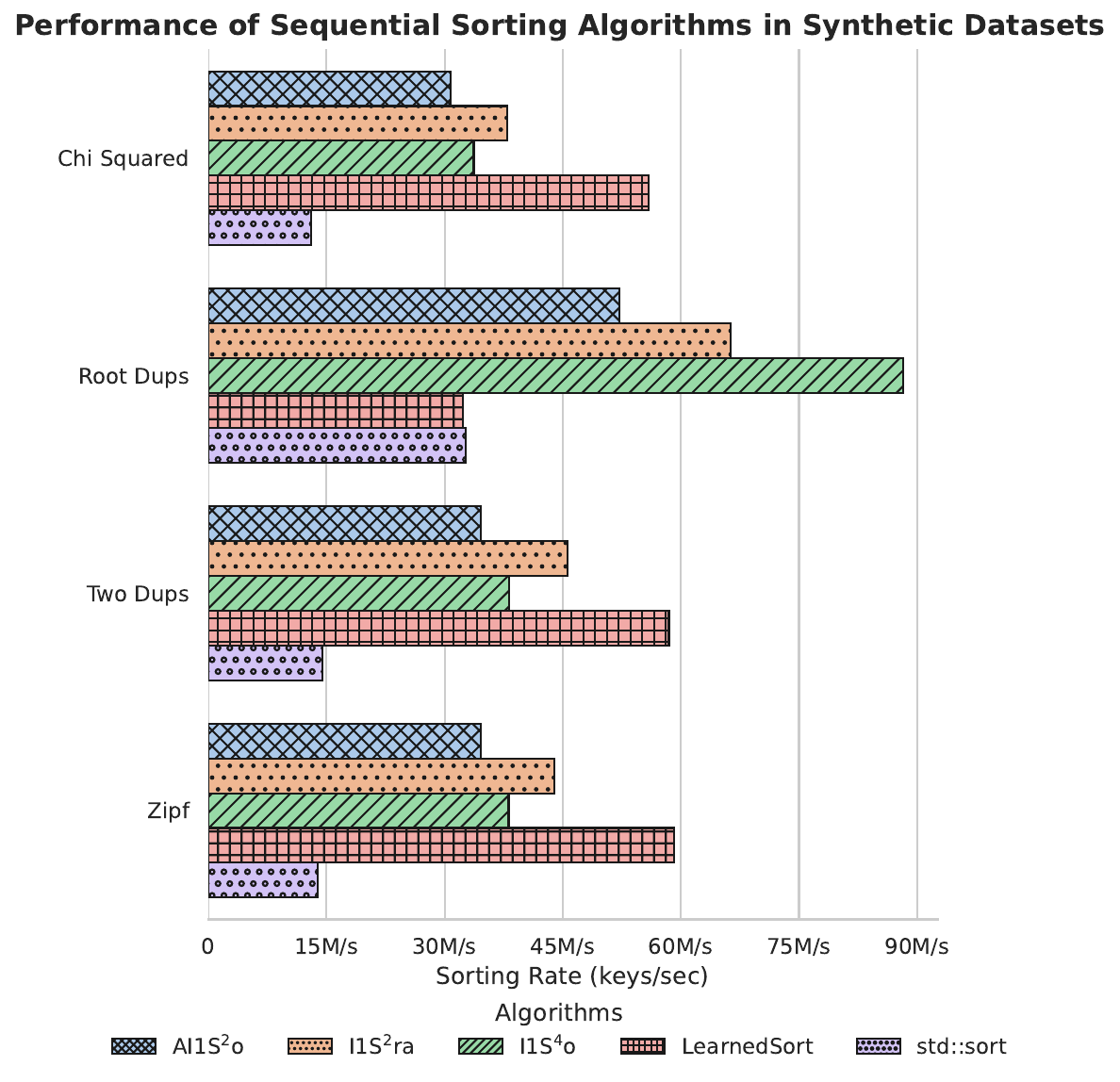}
    \caption{Sorting throughput of the sequential algorithms. Higher rates are better.}
    \label{fig:seq_synth_b}
\end{figure}

\begin{figure}[!htbp]
    \includegraphics[width=\linewidth]{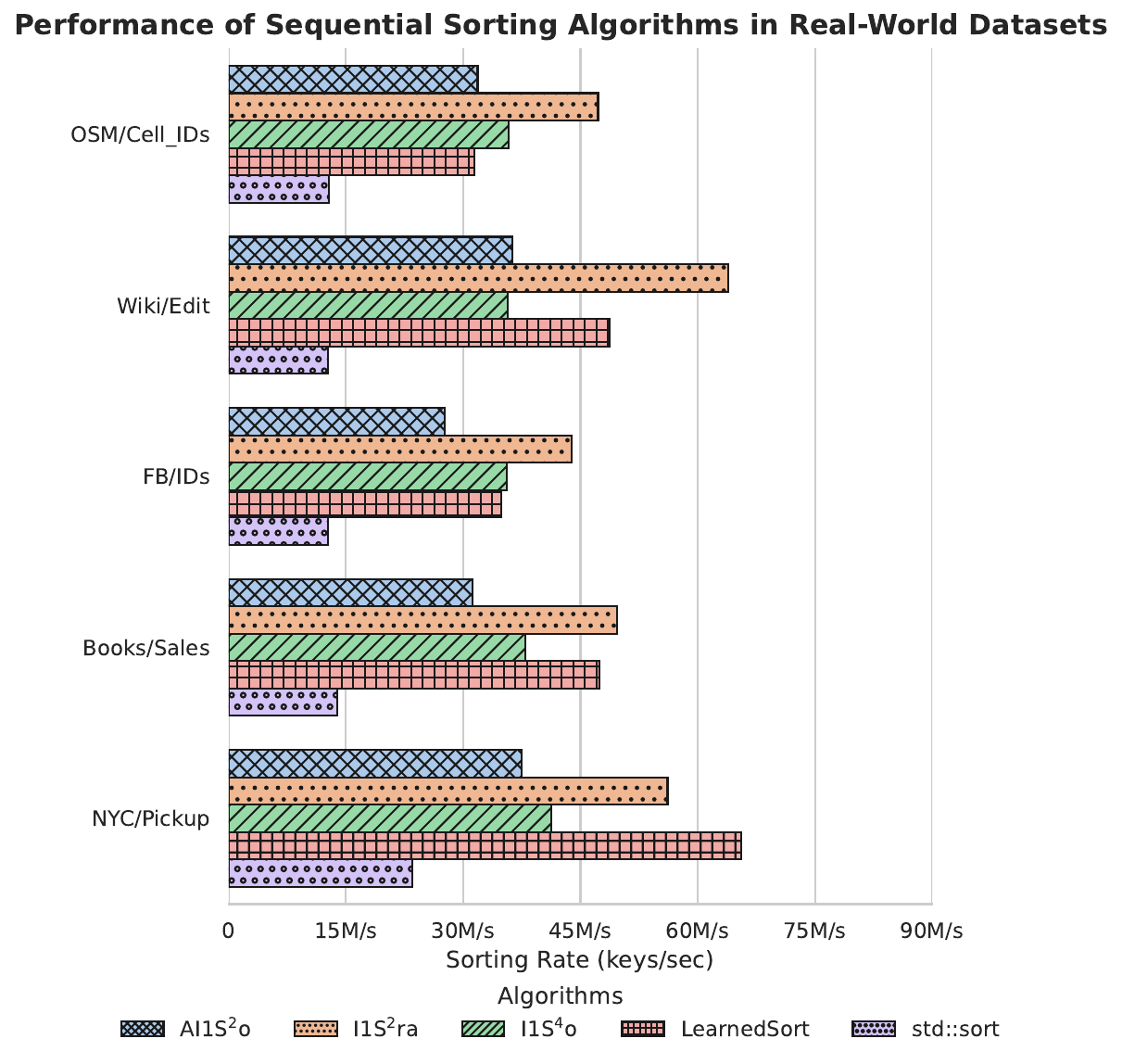}
    \caption{Sorting throughput of the sequential algorithms. Higher rates are better.}
    \label{fig:seq_real}
\end{figure}

The sorting rate of the sequential algorithms is in Figures \ref{fig:seq_synth_a}, \ref{fig:seq_synth_b}, and \ref{fig:seq_real}. The rate is measured by keys per second and indicates the throughput of each algorithm. The numbers are the mean of 10 executions of the algorithms. Higher rates indicate better algorithms.

LearnedSort is the fastest in 9 out of 14 of the datasets, claiming the first spot in the sequential benchmark. \(\text{I1S}^{2}\text{Ra}\) comes second, beating the competitors in 4 datasets. Surprisingly, \(\text{I1S}^{2}\text{Ra}\) outperforms LearnedSort in most of the real-world datasets that were created to benchmark the RMIs that power LearnedSort. \(\text{I1S}^{4}\text{o}\) is the fastest only for one dataset, Root Dups, that it handles gracefully due to its equality buckets.

\(\text{AI1S}^{2}\text{o}\) is outperformed in the sequential benchmark. It is faster than the baseline of \texttt{std::sort}. Nevertheless, the hybrid algorithm is slower than both LearnedSort and \(\text{I1S}^{4}\text{o}\) that provide its inner parts.

We attribute the slower sequential results to the more costly training step of \(\text{AI1S}^{2}\text{o}\). It is important to recall that the training time is accounted in the sorting time for \(\text{AI1S}^{2}\text{o}\) and LearnedSort. \(\text{AI1S}^{2}\text{o}\) samples more data than \(\text{I1S}^{4}\text{o}\) on each partitioning step, which incurs a penalty as we need to sort those samples. The advantage of having better pivots is offset by the training cost. \(\text{AI1S}^{2}\text{o}\) also spends more time training models than LearnedSort as LearnedSort trains the RMI only once while \(\text{AI1S}^{2}\text{o}\) trains a RMI per recursive call.

As we will see in the next section, \(\text{AIPS}^{2}\text{o}\) is a more competitive parallel algorithm. We found that adjusting the sample size and training time had little to no improvement on the sequential case but improved the parallel performance.

\subsection{Parallel Results}

\begin{figure}[!htbp]
    \includegraphics[width=\linewidth]{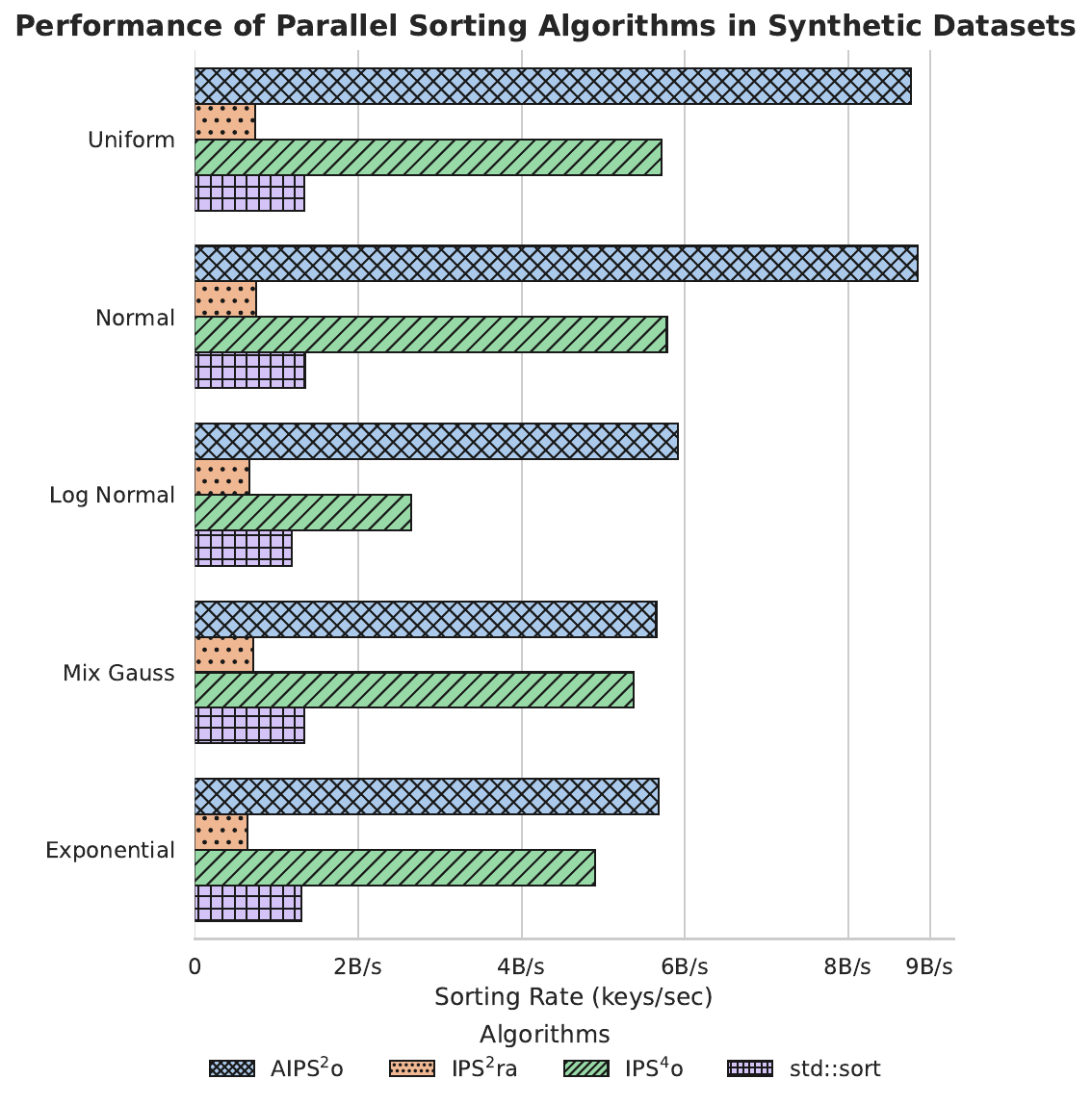}
    \caption{Sorting throughput of the parallel algorithms. Higher rates are better.}
    \label{fig:par_synth_a}
\end{figure}

\begin{figure}[!htbp]
    \includegraphics[width=\linewidth]{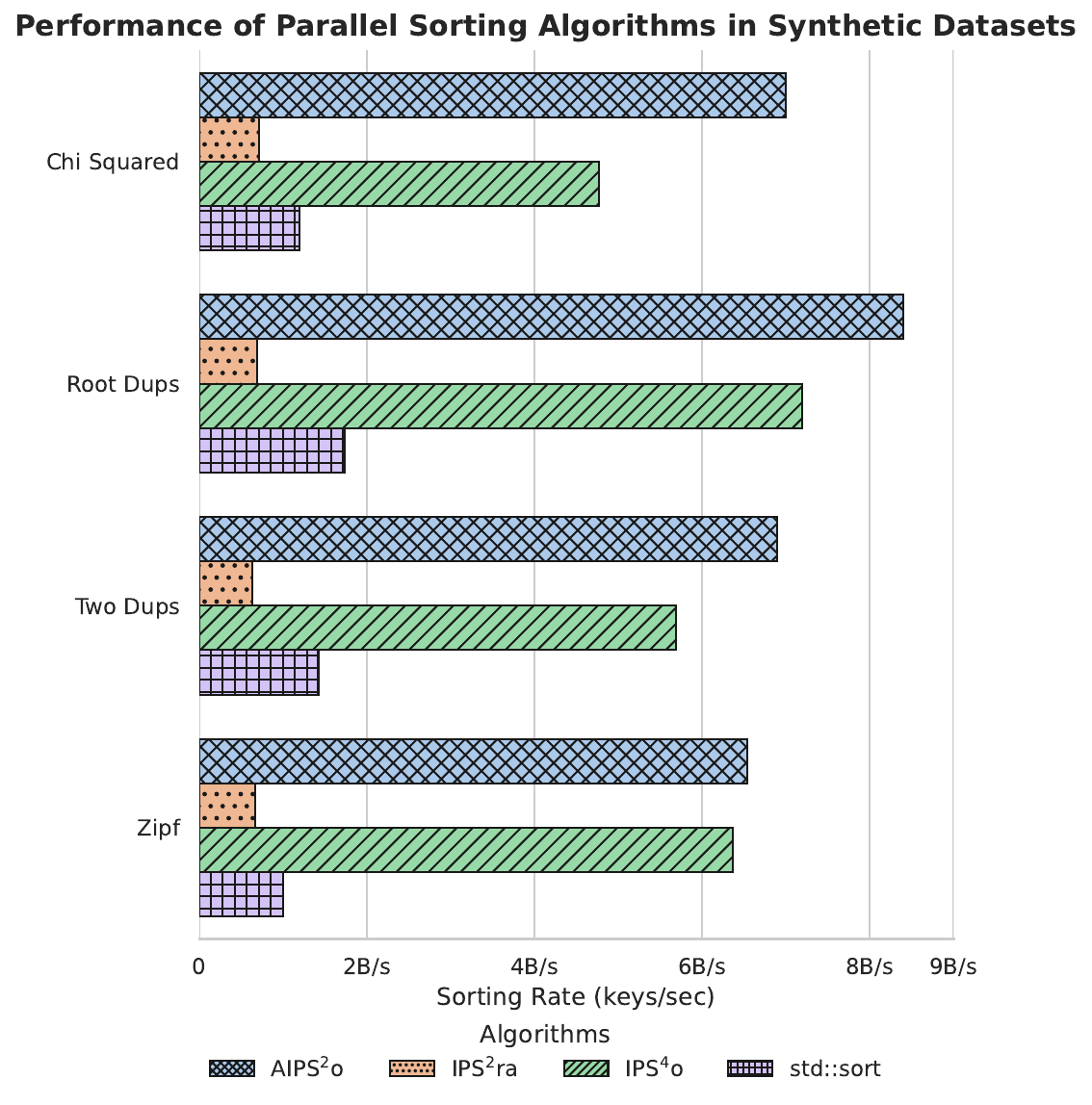}
    \caption{Sorting throughput of the parallel algorithms. Higher rates are better.}
    \label{fig:par_synth_b}
\end{figure}

\begin{figure}[!htbp]
    \includegraphics[width=\linewidth]{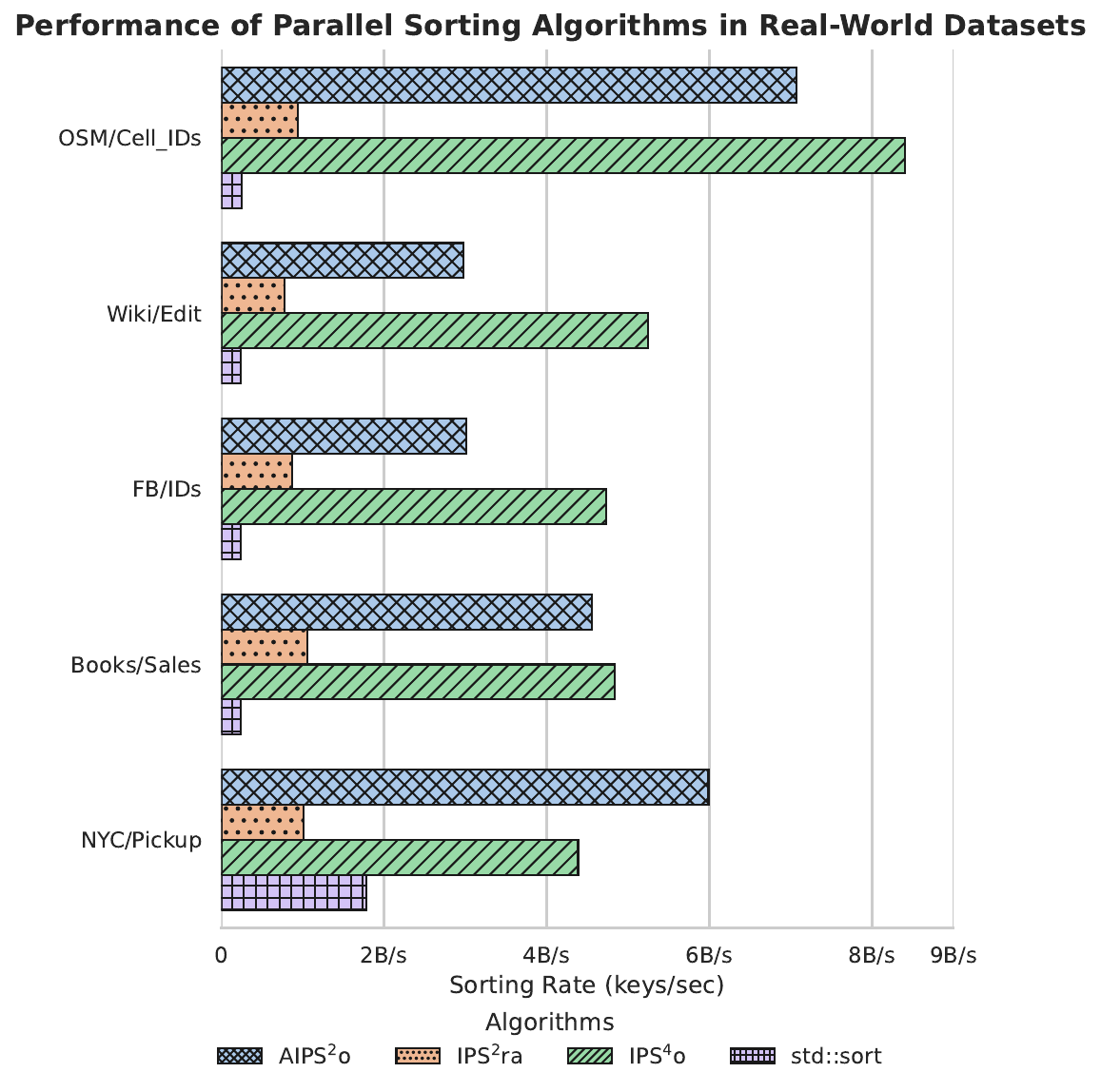}
    \caption{Sorting throughput of the parallel algorithms. Higher rates are better.}
    \label{fig:par_real}
\end{figure}

The sorting rate of the parallel algorithms is in Figures \ref{fig:par_synth_a}, \ref{fig:par_synth_b}, and \ref{fig:par_real}. The rate is measured by keys per second and indicates the throughput of each algorithm. The rates come from the mean of 10 executions of the algorithms.

\(\text{AIPS}^{2}\text{o}\) is the fastest in 10 out of 14 of the datasets, claiming the first spot in the parallel benchmark. \(\text{IPS}^{4}\text{o}\) comes second finishing as the fastest in 4 out of 14 datasets. \texttt{std::sort} places third. \(\text{IPS}^{2}\text{Ra}\) finishes last, behind the baseline on the majority of cases.

The key to high parallel performance is an algorithm's abilitity to maximize the use of the hardware. \(\text{AIPS}^{2}\text{o}\)  creates the best partition of the data for the majority of cases, which creates many sub-problems of a balanced size. This favours the performance of \(\text{AIPS}^{2}\text{o}\) because it manages to keep every thread of the CPU busy always doing work. It also hurts \(\text{AIPS}^{2}\text{o}\) when the RMI does not model the data as accurately. The lowest throughputs of \(\text{AIPS}^{2}\text{o}\) happen on the FB/IDs and Wiki/Edit datasets, which are known to be harder for the RMI than the Books/Sales and OSM/Cell\_IDs datasets \cite{Maltry2022}.

By contrast, \(\text{IPS}^{2}\text{Ra}\) does not manage to use all the hardware because its partitions are not balanced. There are no bounds on the number of elements that have the same radix prefix and go in the same bucket. Hence, \(\text{IPS}^{2}\text{Ra}\) may end with threads waiting for work, hurting its sorting rate compared to \(\text{AIPS}^{2}\text{o}\) and \(\text{IPS}^{4}\text{o}\) which always keep threads busy. This is particularly relevant to show that having a fast sequential algorithm does not necessarily imply a fast parallel algorithm and vice-versa.

The benchmarks demonstrate that \(\text{AIPS}^{2}\text{o}\) is a practical algorithm. It is a parallel LearnedSort that achieves excellent sorting rates in many datasets. We expect that continuous work will help \(\text{AIPS}^{2}\text{o}\) become more robust against data distributions like the one from FB/IDs, finally closing the gap between \(\text{AIPS}^{2}\text{o}\) and \(\text{IPS}^{4}\text{o}\) on the cases where the latter wins.

\section{Conclusion and Future Work}

This paper argues that LearnedSort is analogous to a SampleSort with pivots selected by a CDF model. This helps explain the effectiveness of LearnedSort by comparing it to SampleSort. We introduced the Augmented In-place Parallel SampleSort, combining the state-of-the-art implementation of SampleSort with LearnedSort. The benchmarks demonstrated that Augmented In-place Parallel SampleSort is a practical parallel implementation of LearnedSort that can outperform the fastest parallel sorting algorithm in the majority of the tested inputs including both synthetic and real-world data sets.

Future work in this research direction is to explore how machine learning models can be applied to other use cases in sorting. Some possibilities include:

\textbf{GPU Sorting}: Can the RMI or other learned indexes be combined with GPU SampleSort \cite{Leischner2010}?

\textbf{String Sorting}: Can learned indexes targeting strings \cite{Spector2021} be combined with String SampleSort \cite{Bingmann2013}?

\textbf{Sampling and Pivot Quality}: Can the quality of the learned pivots improve if combined with better sampling techniques  \cite{Harsh2019}?

\balance

\bibliographystyle{ACM-Reference-Format}
\bibliography{refs}

\end{document}